# Android Botnet Detection using Convolutional Neural Networks

Sina Hojjatinia[1], Sajad Hamzenejadi[2], and Hadis Mohseni[3]

[1, 2, 3] *Department of Computer Engineering, Shahid Bahonar University of Kerman*
[1, 2] {sinahojjatinia, sajadhamzenejadi76}@gmail.com
[3] hmohseni@uk.ac.ir

*Abstract*— Today, Android devices are able to provide various services. They support applications for different purposes such as entertainment, business, health, education, and banking services. Because of the functionality and popularity of Android devices as well as the open-source policy of Android OS, they have become a suitable target for attackers. Android Botnet is one of the most dangerous malwares because an attacker called Botmaster can control that remotely to perform destructive attacks. A number of researchers have used different well-known Machine Learning (ML) methods to recognize Android Botnets from benign applications. However, these conventional methods are not able to detect new sophisticated Android Botnets. In this paper, we propose a novel method based on Android permissions and Convolutional Neural Networks (CNNs) to classify Botnets and benign Android applications. Being the first developed method that uses CNNs for this aim, we also proposed a novel method to represent each application as an image which is constructed based on the co-occurrence of used permissions in that application. The proposed CNN is a binary classifier that is trained using these images. Evaluating the proposed method on 5450 Android applications consist of Botnet and benign samples, the obtained results show the accuracy of 97.2% and recall of 96% which is a promising result just using Android permissions.

## I. INTRODUCTION

Compared to conventional mobile phones which mainly provide services for messaging and voice calls, smartphones are general-purpose devices with high computing power and support applications for entertainment, Web browsing, social networking, banking services, etc.

Smartphones need an Operating System (OS) to manage the device, as well as provide an environment for running various applications and providing diverse services for users. There are several operating systems such as Android, iOS, Windows Phone, Symbian, and Blackberry which are used by Smartphones. Among them, Android and iOS are used more, as reported by International Data Corporation (IDC) website [1], Android dominated the smartphone market by a share of 87.0%, followed by iOS with a share of 13.0%, leaving an insignificant market share for the other platforms.

Based on the growing popularity of mobile devices and Android OS, Android-based mobile devices have become a suitable target for attackers. Moreover, because of Android's policy of the open-source kernel, attackers can analyze the Android OS in detail to design sophisticated malware [2], [3]. Besides, as mobile devices are often connected to the Internet to receive online services, attackers have the opportunity to control mobile devices using the Internet. According to the McAfee Lab's 2018 threat report [4], the number of new mobile malware has increased to 2,000,000 until Q4 2018. According to the above information, it is necessary to increase the security of mobile devices using more innovative methods.

There are different types of mobile malware including Mobile Botnet, Worms, Trojans, Viruses, etc. [5]. As a very dangerous type, this paper focuses on mobile Botnet, which is defined as a collection of mobile devices infected with a malicious application or program called Bot, and controlled remotely by an attacker called Botmaser. The Botmaster controls the infected mobile devices using Command and Control (C&C) channel to perform illegal activities such as to send premium-rate messages, steal user information, DDoS attack, email spamming, phishing, etc. [2]. During an attack, mobile users may not perceive that their mobile is a part of a Botnet since the mobile bots operate silently and they usually show no suspicious behavior. For instance, DroidDream bot acquire root privilege to prevent from being deleted and installs additional malicious programs when the user mobile is asleep [6]. As another example, RootSmart bot uses default Android application icons to evade being detected by the user before acquiring root privilege [7].

Therefore, according to the provided explanation, there is an urgent need to design new methods to overcome mobile Botnet. Because of the popularity of Android mobile devices, the purpose of this paper is to propose an innovative method to detect Botnets in Android-based devices. This is done by the help of training a deep network using Android permissions used in mobile applications as informative input features. These features are extracted by analyzing a collection of benign and malicious applications from the Google play store and the ISCX dataset [8]. Proposing a novel method to present the extracted features through black and white images, a convolutional neural network (CNN) is trained to detect Botnet applications from benign ones. Many previously developed approaches use traditional machine learning (ML) methods to detect Android Botnets. However, to the best of our knowledge, this is the first approach that uses CNNs to detect Android Botnets.

The rest of the paper is organized as follows. Section II is devoted to an overview of the related works. In Section III, after providing background about Android Botnets and CNNs, the proposed method is presented in detail. Section IV evaluates the proposed method and discusses experimental results. Finally, Section V concludes the paper.

## II. Related Works

In recent years, the field of Android Botnet detection has appealed the attention of many researchers. Generally, Android Botnet detection methods can be divided into two separate groups namely Signature-based detection, and Anomaly-based detection [9]. The signature-based detection systems detect intrusions by comparing the behavior of mobile devices with the signatures or patterns of malicious applications. If the behavior of the mobile device seems similar to malicious patterns, they generate an alert. Signature-based methods can detect known Android Botnets with high accuracy. However, they are unable to detect new attacks that are still undefined for them. In other words, these methods need a database of malicious patterns and this database should be updated constantly. On the other hand, Anomaly-based methods try to solve the drawbacks of signature-based methods by learning the normal behavior of a mobile device using learning methods. Any behavior that is not learned as a normal one is considered as an anomaly.

A number of authors [10-12] have used signature-based methods to detect Android Botnets. Alzahrani et al. [10] proposed a real-time method to detect SMS Botnets in mobile devices. They analyzed the content of SMS messages such as URLs, specific words, and phone number to detect SMS Botnets. The log analysis is used by Girei et al. [11] to detect Android Botnets by proposing the Logdog architecture. The authors used logcat, which is the Android logging system to collect Android logs and send them to a cloud-based system for further analysis. Yusof et al. [12] used Android API calls and permissions to propose a GPS-based Botnet classification. They found 29 patterns of permissions and API calls that use GPS functionality and reported that these 29 patterns can effectively classify Android Botnets that exploiting the GPS service. However, all these mentioned methods suffer from the major drawbacks of signature-based Botnet detection.

Pushing the limitations of signature-based methods, a number of authors [13-17] used anomaly-based methods to detect Android Botnets. Jadhav et al. [13] proposed a multi-layer approach to classify the families of Android Botnets using features such as system calls, network traffic, and application level function call data. The proposed system performs based on the cloud and users have to send applications for security analysis. This system sends back a brief security analysis report to the user to indicate the application is benign or not. However, their system requires a JAVA application to be installed on the user's mobile device. MBotCS proposed by Meng et al. [14] detects mobile Botnets using network features such as TCP/UDP packet size, frame duration, and source/destination IP address. The authors used a set of ML box algorithms and five different ML classifiers to classify the collected network traffic in real-time. However, their method suffers from high false positive rate. Alqatawna et al. [15] used Android permissions and their corresponding protection levels to detect Android Botnets. The authors collected their samples from the Google play store and ISCX dataset and tested their method using four well-known ML classifiers. Yusof et al. [16] presented a method to classify Android Botnets by using API calls and permissions. They extracted 31 API calls and 16 permissions from 10 different families of Android Botnets and classified their samples using five different ML classifiers. However, the drawback of their method is the high rate of false positive. A distinguished method proposed by Mongkolluksamee et al. [17] where they used a specific graph called graphlet to capture communication patterns for the aim of detecting peer-to-peer mobile Botnets. The authors generated their dataset by capturing the traffic of a sophisticated mobile Botnet, called NotCompatible.C, over 5 mobile devices. Also, principal component analysis (PCA) was used by the authors to increase the performance of their method. However, they only used Random Forest algorithm as their classifier and evaluated their method using only F-Measure. Other metrics such as precision, accuracy, and FPR were not reported by them.

Based on the above-mentioned researches, one can conclude that traditional ML algorithms are used widely for the Android Botnet detection aim. However, the contribution of our proposed method lies in investigating the idea of applying deep learning for Android Botnet detection, while the input data has been presented in a novel image-based manner.

## III. Android Botnet Detection

### A. Background

As a trend in artificial intelligence, deep learning (DL) is used to solve various complicated problems such as speech recognition [18], computer vision [19], autonomous driving [20], and natural language processing (NLP) [21]. DL is a branch of ML that learns from representing data in abstract concepts within hierarchical trainable layers. DL models are faster and more accurate than traditional ML algorithms. Besides, they are able to manipulate large datasets [22]. As a subset of DL, convolutional neural networks (CNNs) mainly solve image and video recognition problems. CNN is a feed-forward neural network and its hidden layers comprise three distinct operational layers as in the following.

1. *Convolutional layer*, which is the most significant block of CNN architecture. It is responsible for extracting high-level features by convolving a specified number of kernels on the input image.
2. *Pooling layer* is an optional block of CNN architecture that reduces the dimensionality of input mapped features without significant information loss. It aims to reduce computational cost and control the overfitting problem.
3. *Fully Connected layer* is a dense layer that finalizes the classification task after learning features in the previous layers.

In the following, we first describe each application by constructing an image from its permissions using a novel method. Then, since CNNs are state-of-the-art methods for image recognition [22], we propose our CNN model to detect Android Botnet applications using the constructed images.

### B. Proposed Method

Android Package Kit (apk) is the necessary part of each application to be executable on Android OS. The apk file is a compressed file that mainly contains three components namely resource files, code files, and *AndroidManifest.xml* file. Among them, *AndroidManifest.xml* file contains information such as requested permissions, broadcast receivers, intents, and permission protection levels. Permissions are the requests which asked by applications during the installation process to access to mobile resources such as memory, GPS, camera, internet, and Wi-Fi. However, applications may not request permissions for providing legal services. Fig. 1 and Fig. 2 show the top 15 Android permissions used by Botnet and benign applications, respectively. The figures depict that there are two different orders of frequently used permissions used by Botnet and benign applications. For example, SEND_SMS is ranked 14th and 5th among the whole permissions used by benign and Botnet applications, respectively. Also, comparing the two figures, one can infer that Botnet applications use more permissions compared to benign ones. For example, 90.48% of Botnet applications use READ_PHONE_STATE where only 33.08% of benign applications use this permission. Obviously, using more permissions, Botmasters are able to perform operations that are more dangerous. Moreover, most of the mobile users do not care about the permissions asked by applications, which encourages the Botmasters to obtain further access by asking more permissions. Inspired by these reasons, the proposed method of this paper uses Android permissions as describing features of each application and utilizes these features to represent each application as an image.

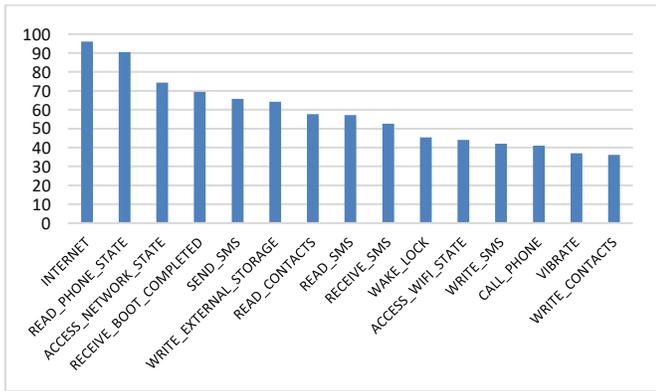

Fig. 1. Top 15 permissions used by Botnet Android applications

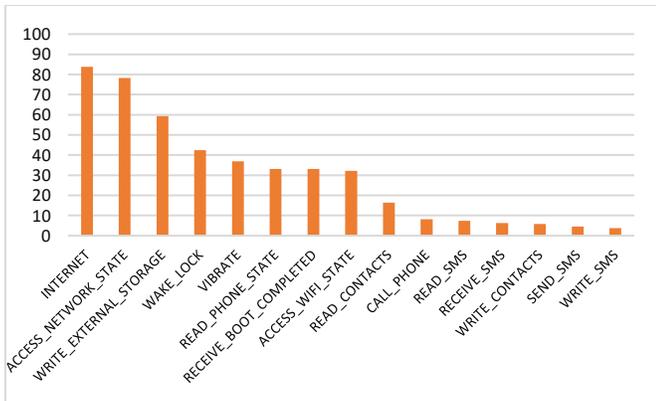

Fig. 2. Top 15 permissions used by benign Android applications

*1) apk Pre-processing*

In the first part of the method, we processed a collection of apk files to achieve two sets of permissions that will be used later in the application representation process. This part includes four main steps:

1. A total number of 5450 apk files (3650 benign and 1800 Botnet applications) were collected to build our dataset.
2. Using reverse engineering on apk files, *AndroidManifest.xml* files were extracted to determine the requested permissions by each application. This was done with the help of Apktool [23], which is a popular open-source tool for reverse engineering.
3. A tool was developed using python to extract permissions of each application from its *AndroidManifest.xml* file. Creating two distinct lists for Botnet and benign applications, the extracted permissions from each application were appended to its corresponding list.
4. Each list was sorted based on the most frequently used permissions.

*2) Representing Applications as Image*

In the previous subsection, we showed how Android apk files were processed to generate two sorted lists of frequently used permissions. This subsection explains how each application can be presented as an image based on the lists.

For image representation aim, the two mentioned lists were merged in a new sorted list. (the order of permissions in the new list may differ from the original lists). Assuming that there are $n$ permissions in the final list, each application is presented by a $n \times n$ matrix. In this matrix, the $[i,j]$ element shows the co-occurrence of the $i^{th}$ and the $j^{th}$ permissions in an application, if both permissions are used by the application, the $[i,j]$ element is set to 0, otherwise, it is set to 255. Obviously, the diagonal $[i,i]$ element shows if the $i^{th}$ permission in the final merged list is used by the application or not.

Here, we collected the top 41 frequently used permissions in both Botnet and benign applications and represented each application as a $41 \times 41$ black & white image. Fig. 3, depicts some samples from generated images for Botnet and benign applications.

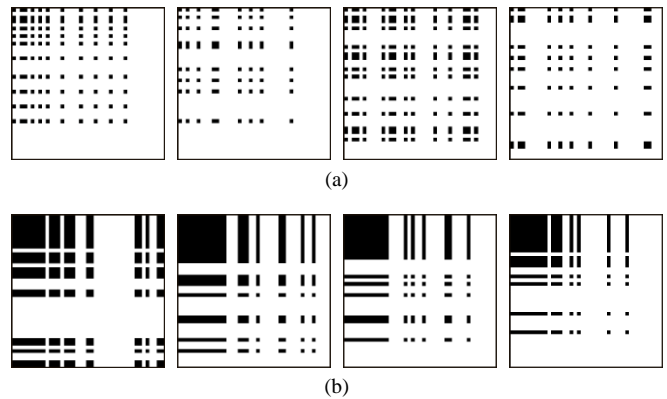

Fig. 3. Different samples of matrix representation from (a) benign and (b) Botnet Android applications. These images indicate that Botnet applications use more permissions than benign applications.

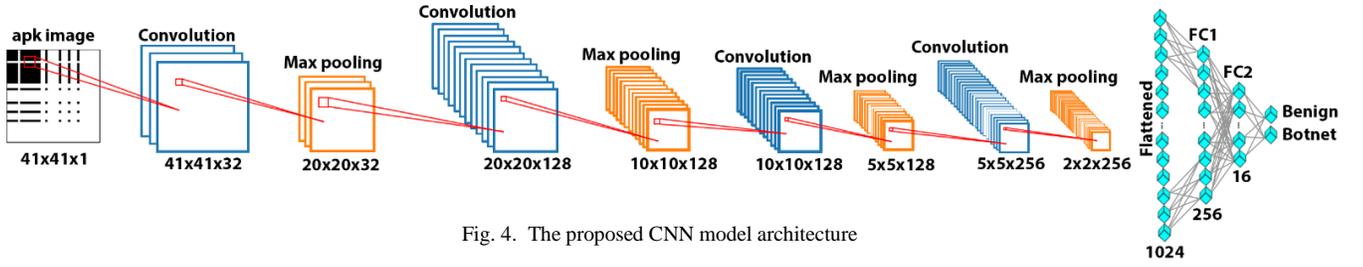

Fig. 4. The proposed CNN model architecture

Table 1. The proposed CNN model configuration

| Layer | Input tensor size | Type | Activation Function | Kernel size | Strides | Kernels | Output tensor size |
|---|---|---|---|---|---|---|---|
| 1 | (41, 41, 1) | Conv | ReLU | 5 × 5 | (1, 1) | 32 | (41, 41, 32) |
| 2 | (41, 41, 32) | Max-pool | - | 2 × 2 | (2, 2) | - | (20, 20, 32) |
| 3 | (20, 20, 32) | Conv | ReLU | 5 × 5 | (1, 1) | 128 | (20, 20, 128) |
| 4 | (20, 20, 128) | Max-pool | - | 2 × 2 | (2, 2) | - | (10, 10, 128) |
| 5 | (10, 10, 128) | Conv | ReLU | 3 × 3 | (1, 1) | 128 | (10, 10, 128) |
| 6 | (10, 10, 128) | Max-pool | - | 2 × 2 | (2, 2) | - | (5, 5, 128) |
| 7 | (5, 5, 128) | Conv | ReLU | 1 × 1 | (1, 1) | 256 | (5, 5, 128) |
| 8 | (5, 5, 256) | Max-pool | - | 2 × 2 | (2, 2) | - | (2, 2, 256) |
| 9 | (2, 2, 256) | FC-1 | ReLU | - | - | - | (256, 1) |
| 10 | (256, 1) | FC-2 | ReLU | - | - | - | (16, 1) |
| 11 | (16, 1) | Softmax | - | - | - | - | (2, 1) |

*3) CNN Configuration*

As the final part of the proposed method, a CNN is trained to distinguish Botnet applications from benign ones. Table 1 shows the CNN configuration of the model used by the proposed method. Moreover, Fig. 4 illustrates the architecture of the CNN model. The model uses Adam optimizer algorithm [24] and since it is trained for a two-class problem, binary cross-entropy [25] is used as the loss function, which is defined as follows.

$$H_p(q) = -\frac{1}{N}\sum_{i=1}^{N} y_i \cdot \log(p(y_i)) + (1 - y_i) \cdot \log(1 - p(y_i)) \quad (1)$$

Here, N is the number of samples and y is a binary number that specifies the label of sample (i.e. Botnet or benign).

IV. EXPERIMENTS

This section explains the done experiments and provides a comparison between the proposed method and the most related previous researches in this filed.

*A. Dataset and Experiment Settings*

For evaluating the proposed method, we used the ISCX [8] dataset and selected 1800 Android Botnet samples from 14 different families. For collecting the normal samples, we developed a tool to crawl the Google play store and downloaded totally 3650 samples from 24 different categories. All the benign samples scanned with VirusTotal project [26] to make sure that the benign class does not include any malware sample. Totally, the dataset consists of 5450 samples and 10-fold cross-validation was used to evaluate the proposed method using this dataset.

All experiments were performed on 64-bits Ubuntu 18.04 operating system and using a hardware with Intel(R) Core(TM) i5-6200u @ 2.30GHz CPU, 8GB DDR4 RAM, and an NVIDIA GeForce 920M GPU with 2GB memory. More specifically, the proposed method was implemented using TensorFlow framework [27] by Keras library [28].

*B. Experimental Results*

Assuming Botnet and benign samples as TRUE and FALSE classes, respectively, five different metrics consist of false positive rate (FPR) (2), Recall (3), Precision (4), Accuracy (5), and F-Measure (6) were used to evaluate the performance of the proposed method. These metrics are defined as follows.

$$\text{FPR} = \frac{FP}{FP + TN} \quad (2)$$

$$\text{Recall} = \frac{TP}{TP + FN} \quad (3)$$

$$\text{Pricision} = \frac{TP}{TP + FP} \quad (4)$$

$$\text{Accuracy} = \frac{TP + TN}{TP + TN + FP + FN} \quad (5)$$

$$\text{F\_Measure} = \frac{2 \times \text{Precision} \times \text{Recall}}{\text{Pricision} + \text{Recall}} \quad (6)$$

Where FP, TN, TP, and FN stand for False Positive, True Negative, True Positive, and False Negative, respectively.

Table 2. A comparison between our method and most related works

| Reference | ML/DL method | Dataset | Features | Performance (%) | | | | |
|---|---|---|---|---|---|---|---|---|
| | | | | ACC | R | P | F-Measure | FPR |
| **Our method** | **CNN** | **3650 benign samples from Google play and 1800 Botnet samples from ISCX dataset** | **Permissions** | **97.2** | **96** | **95.5** | **95.7** | **2.2** |
| [15] | Random Forest, MLP, Decision Tree, Naïve Bayes | 1635 normal samples from Google play and 1635 botnet samples from ISCX dataset | Permissions and its corresponding protection level | 97.3 | 95.7 | 98.7 | - | - |
| [29] | SVM, Random Forest, Bagging, Naïve Bayes, Decision Tree, MLP, SMO | 150 normal samples from Google play and 1926 botnet samples from ISCX dataset | API Calls and Permissions | - | 96.9 | 97.2 | - | - |
| [30] | SVM, kNN, Decision Tree(J48), Bagging, Naïve Bayes, Random Forest | 1400 normal samples from 1400 botnet samples from ISCX dataset | Permissions | 95.1 | 82.7 | 97 | - | - |
| [31] | Naïve Bayes, Decision Tree, Random Forest | 850 normal samples from Google play and 1505 botnet samples from ISCX dataset | Permissions | - | 94.6 | 93.1 | - | 9.9 |
| **ACC refers to Accuracy, R refers to Recall, P refers to Precision, FPR refers to False Positive Rate** | | | | | | | | |

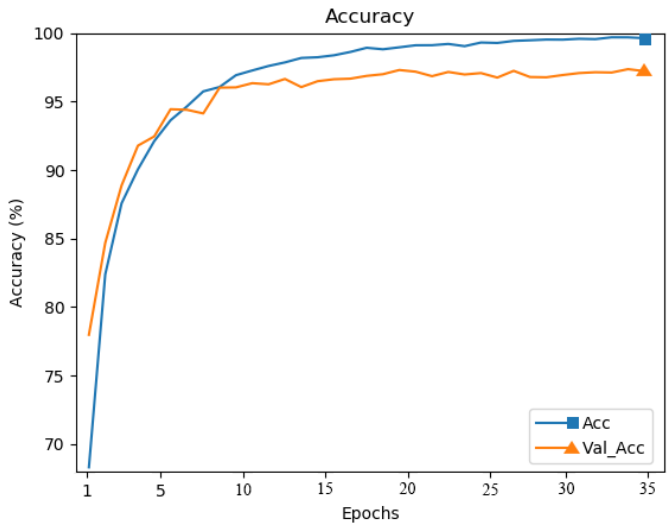

Fig. 5. Variations of accuracy over 35 epochs of network training.

We trained and tested the proposed CNN model in 10, 15, 20, 25, 30, and 35 epochs as the plot in Fig. 5 shows. Based on this figure, there is no significant improvement in both training and validation accuracy after 25 epochs and training the proposed network with 25 epochs shows the best performance in an average of 10-fold cross validation with the accuracy of 97.2% and recall of 96%.

Table 2 shows the obtained results from the proposed method compared to other best researches that have used traditional machine learning approaches (in terms of used dataset, number of samples and performance). The table indicates that the proposed method is completely successful in classifying benign and Botnet applications. It should be noted that compared to other methods, we have used the most complete and also complex dataset. It covers many benign samples from Google Play and all the difficult samples from ISCX dataset, and is more compatible with the real-world problem. Moreover, we distinguished Botnet and benign applications just by utilizing permissions while some of the mentioned works used more features alongside permissions such as API calls or permissions protection level. Our promising results with high accuracy and low FPR indicate that our method can effectively detect Android Botnets based on only used permissions as features.

## V. CONCLUSION AND FEATURE WORKS

In this paper, we used Android permissions and convolutional neural networks to propose a novel method to detect Android Botnets. To the best of our knowledge, this is the first Android Botnet detection method that uses convolutional neural networks. Initially, reverse engineering was used on 3650 benign applications from Google play store and 1800 Botnet applications from ISCX dataset to extract the *AndroidManifest.xml* files that provided access to the used permissions by each application. In the second step, we used permissions and their co-occurrence in an application to represent each application as an image. Finally, we trained and tested a proposed deep convolutional neural network model using the describing images in a 10-fold cross validation experiment. Based on done experiments, the proposed method outperforms several conventional ML methods in this filed by achieving 97.2% accuracy and 96% recall. These promising results show that the proposed method can effectively detect Android Botnets by using permissions.

As the feature work, we aim to use the Android API calls alongside the permissions to generate images that are more specific to detect sophisticated Android Botnets.